\documentclass[letterpaper, 10 pt, conference]{ieeeconf}  

\IEEEoverridecommandlockouts                          

\usepackage{graphics} 
\usepackage{epsfig} 
\usepackage{mathptmx} 
\usepackage{times} 
\usepackage{amsmath} 
\usepackage{amssymb}  
\usepackage{listings}
\usepackage{multirow}
\usepackage{booktabs}
\usepackage{amsfonts}
\usepackage{caption}
\usepackage{lipsum} 
\usepackage{hyperref}
\usepackage{tcolorbox}
\let\labelindent\relax
\usepackage{enumitem}  
\setlist[itemize]{topsep=0pt, left=0pt} 
\usepackage{newtxmath}
\usepackage{arydshln}
\usepackage{float}
\usepackage{units}
\usepackage{todonotes}
\usepackage{sistyle}
\SIthousandsep{,}

\newlength{\oldarrayrulewidth}
\newcommand{\specialhdashline}{\noalign{\vskip 0.2mm}\hdashline\noalign{\vskip 0.5mm}}
\usepackage{tikz}
\newcommand*\circlednum[1]{\tikz[baseline=(char.base)]{
            \node[shape=circle,draw,inner sep=0.1pt] (char) {#1};}}

\newcommand{\modelname}{RACER}
\usepackage{caption}
\usepackage{subcaption}
\title{\LARGE \bf
RACER: Rich Language-Guided Failure Recovery Policies for Imitation Learning}

\author{
Yinpei Dai$^*$$^{1}$, Jayjun Lee$^*$$^{2}$, Nima Fazeli$^{2}$, Joyce Chai$^{1}$
\thanks{$^{1}$Computer Science and Engineering Deparment, $^{2}$Robotics Department, University of Michigan, MI, USA. $^*$ denotes equal contribution.}
\thanks{Emails: \texttt{\{daiyp,jayjun,nfz,chaijy\}@umich.edu}}
\thanks{This work is supported in part by NSF
IIS1949634, NSF SES-2128623, and has benefited from the Microsoft Accelerate Foundation
Models Research (AFMR) grant program.}
}

\begin{document}

\maketitle
\thispagestyle{empty}
\pagestyle{empty}

\begin{abstract}
Developing robust and correctable visuomotor policies for robotic manipulation is challenging due to the lack of self-recovery mechanisms from failures and the limitations of simple language instructions in guiding robot actions. To address these issues, we propose a scalable data generation pipeline that automatically augments expert demonstrations with failure recovery trajectories and fine-grained language annotations for training. We then introduce Rich languAge-guided failure reCovERy (\modelname), a supervisor-actor framework, which combines failure recovery data with rich language descriptions to enhance robot control. \modelname\ features a vision-language model (VLM) that acts as an online supervisor, providing detailed language guidance for error correction and task execution, and a language-conditioned visuomotor policy as an actor to predict the next actions. Our experimental results show that \modelname\ outperforms the state-of-the-art Robotic View Transformer (RVT) on RLbench across various evaluation settings, including standard long-horizon tasks, dynamic goal-change tasks and zero-shot unseen tasks, achieving superior performance in both simulated and real world environments. Videos and code are available at: \href{https://rich-language-failure-recovery.github.io/}{https://rich-language-failure-recovery.github.io}.
\end{abstract}


\section{INTRODUCTION}
Building 
reliable multi-task visuomotor policies for object manipulation through imitation learning is a long-standing challenge in robot learning. 
Recent advances in transformer-based architectures for imitation learning have demonstrated its effectiveness in 
6-DoF Cartesian End-Effector (EE) control \cite{peract, act3d, rvt}. 
Despite this progress, current policies still suffer from an inability to self-recover from online failures during inference time \cite{reflect, liu2022robot, hoque2024intervengen}. Such limitation primarily stems from: (1) these models being trained exclusively on successful expert trajectories, without accounting for failures caused by model mispredictions and inevitable compounding errors, and (2) the absence of mechanisms to efficiently intervene and correct mistakes without requiring humans to take over the control via shared autonomy \cite{karamcheti2022lila, liu2024sirius}.

To address these issues, prior works have focused on using human-in-the-loop interactive imitation learning to closely monitor and rectify robot behaviors through online language corrections \cite{cui2023no, olaf, shi2024yell}. However, these approaches impose a significant burden on human operators to manually intervene robot actions at inference time, and require collecting new online data for iterative model improvement.
Moreover, these approaches typically rely on \textit{simple} language instructions to guide robotic manipulation, such as \textit{``move to the left"} and \textit{``pick up the cup"}, which are insufficient for enabling robots to understand failures and take more accurate corrective actions. 
\begin{figure}
    \centering
    \includegraphics[width=\linewidth]{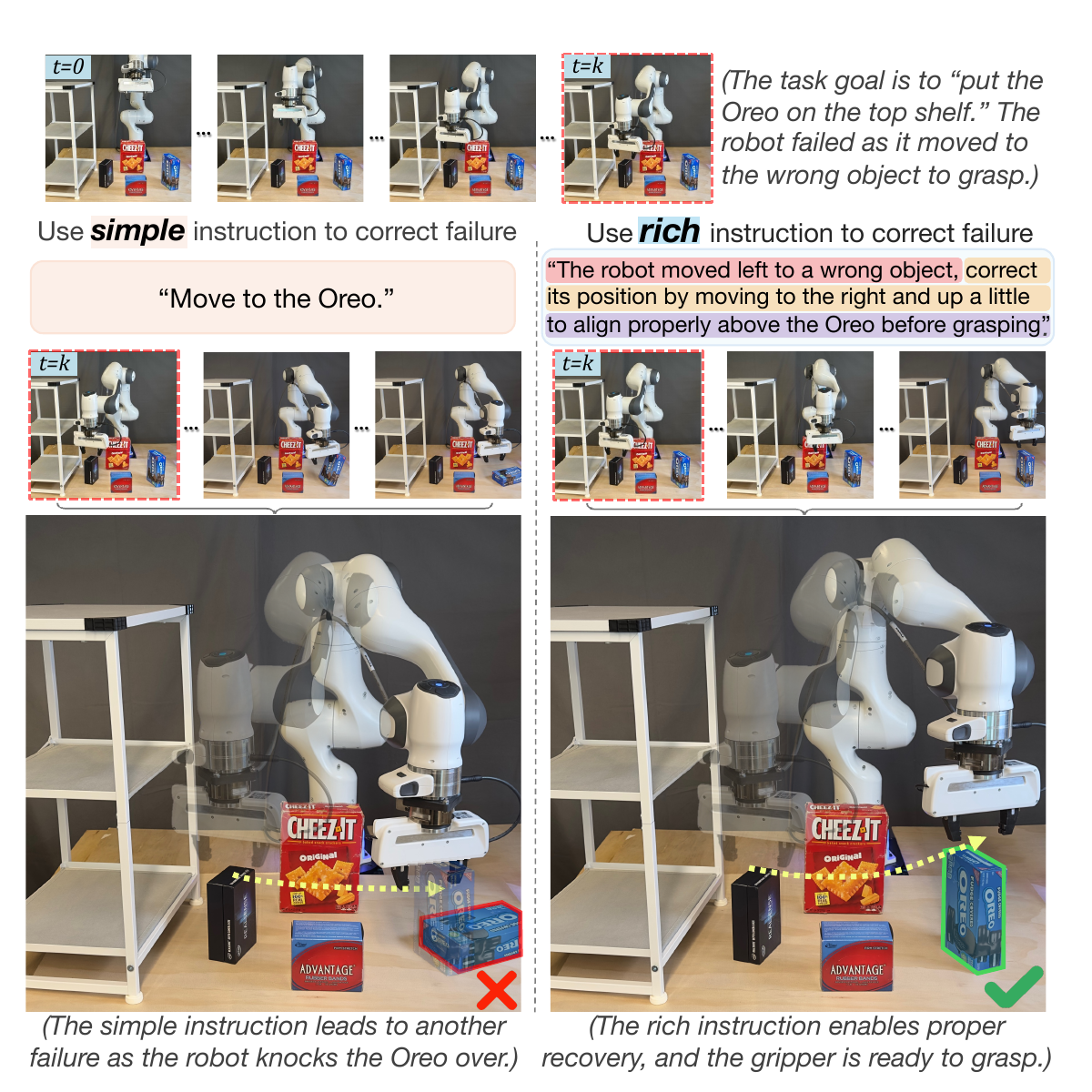}
    \caption{Comparison between the \textit{simple} and \textit{rich} language guidance for failure recovery: The robot should approach the Oreo (the blue box on the right) directly to grasp it but instead moved to the wrong object (the black box).  To help the visuomotor policy recover from this failure, the rich language instruction provides sufficient details, including a \text{\color{red} failure analysis (in red)}, \text{\color{orange} spatial movements (in orange)} and the \text{\color{violet} expected outcome (in purple)}. In contrast, simple language instructions with limited descriptions may not guide the robot effectively, potentially causing it to continue making mistakes.
    }
    \label{fig: teaser}
    \vspace{-20pt}
\end{figure}
We argue that visuomotor policies should learn from failure recovery data paired with \textit{rich} language instructions for effective online recovery. As shown in Fig. \ref{fig: teaser}, to guide the robot to a proper recovery pose, language guidance should not only specify basic actions (e.g., predicates and objects) but also contain more sufficient information about  failure analysis, spatial movement descriptions and the expected outcome of taking the current action.
This richer, more descriptive language support is crucial for enhancing the robot's ability to comprehend complex scenarios, recover from errors, and ultimately improve the overall performance.

However, most existing popular benchmarks lack either failure recovery data or trajectories paired with rich language descriptions \cite{james2020rlbench, mees2022calvin, yu2020metawld, mandlekar2023mimicgen}. Therefore, we propose an automatic failure data augmentation pipeline that extends expert demonstrations from RLBench \cite{james2020rlbench} by using random perturbations to generate failure data and leveraging large language models (LLMs) to annotate rich language descriptions for each transition. We then introduce \modelname, a flexible \textit{supervisor-actor} framework that enhances robotic manipulation through rich language guidance for failure recovery. \modelname\ consists of a vision-language model (VLM) as the \textit{supervisor}, which monitors and provides detailed instructions to analyze, correct, and guide robot actions at each step, and a language-conditioned visuomotor policy as the \textit{actor}, responsible for generating the next appropriate action. Our experiments show that \modelname\ significantly outperforms previous state-of-the-art baselines across 18 RLBench tasks, demonstrating superior robustness and adaptability to task goal online changes, unseen task evaluations, and real world scenarios.
We summarize our contributions as follows:

\begin{itemize}
    \item We are the first to explore the role of rich language guidance in robot manipulation and demonstrate its importance in conjunction with failure recovery for robust control.
    \item We propose a scalable language-guided failure recovery data augmentation strategy and collect \num{10159} new trajectories with rich language instructions on RLbench.
    \item We present \modelname, where a VLM instructs a visuomotor policy with rich language. \modelname\ performs competitively on RLbench,  
    demonstrating robustness to task goal changes and generalizability in handling unseen tasks.
    \item We show that \modelname\ enables fast real-world deployment through few-shot sim-to-real transfer, highlighting the role of rich language guidance in bridging the sim-to-real gap.
    
\end{itemize}

\begin{figure*}[h]
    \centering
    \includegraphics[width=0.95\linewidth]{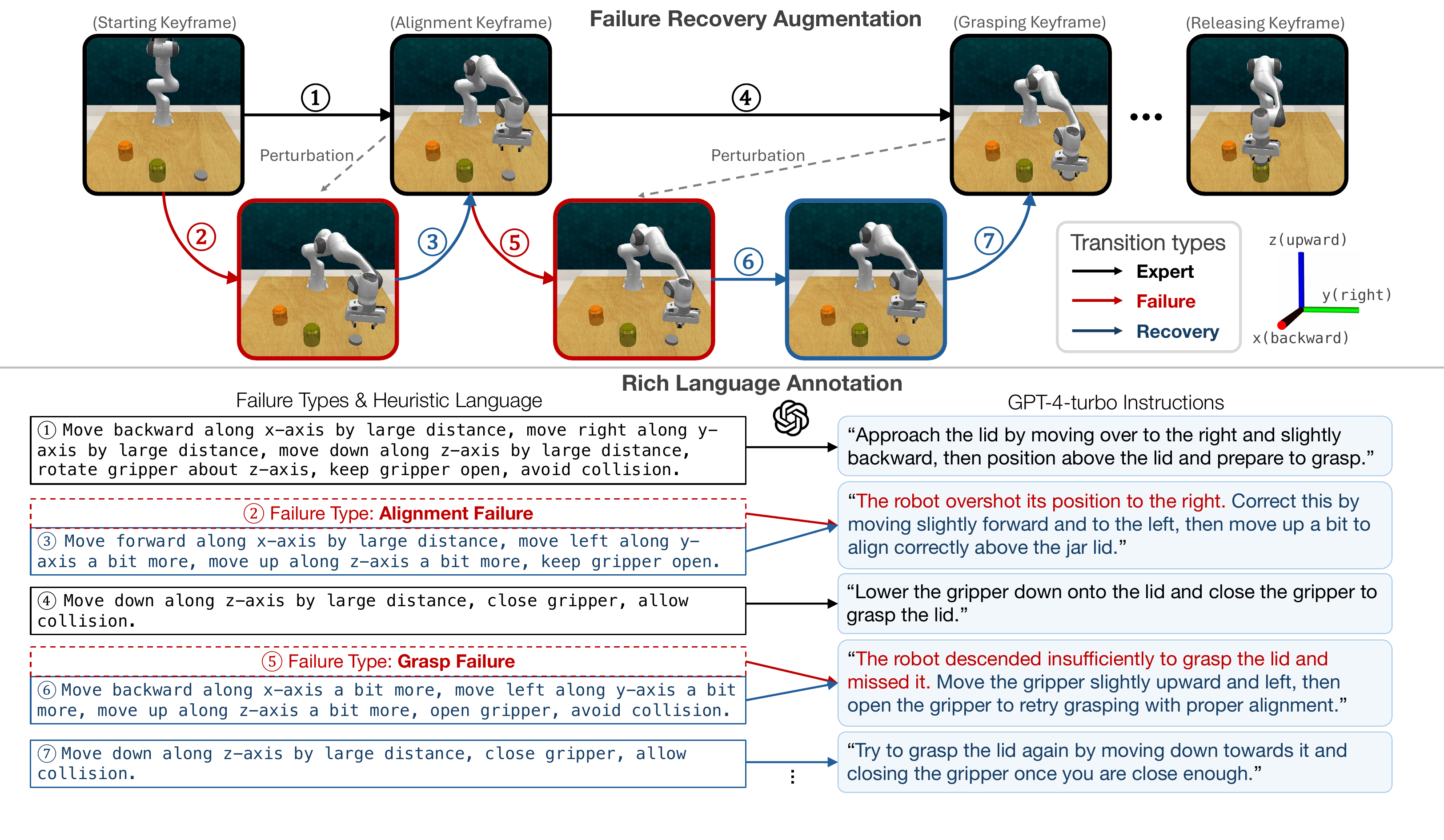}
    \caption{An overview of automatic rich language-annotated failure-recovery data augmentation pipeline. Given an expert demo (e.g., task goal: \textit{close the olive jar}), perturbations are injected to expert actions at crucial keyframes (e.g. aligning to, grasping, and releasing a target object) to induce failures. Then, the expert actions are reused as corrections to collect recovery transitions. Finally, all expert and  recovery transitions are labelled with rich instructions through GPT-4-turbo. The input for GPT-4-turbo includes the task description, ground-truth object locations, failure types, and heuristic language describing the change in the end-effector's pose movement at the current step.}
    \vspace{-15pt}
    \label{fig:data_aug}
\end{figure*}



\section{Related Works}

\textbf{Imitation Learning for Visuomotor Policies.} Imitation learning is commonly used to train visuomotor policies with the supervision of expert demonstrations \cite{10.1145/3054912, zhu2018reinforcement}, predicting actions on either sparse keyframes \cite{peract, arm} or dense waypoints \cite{diffusionpolicy, rt22023arxiv}. 
However, this approach often struggles with out-of-distribution observations, such as failure states. To address this, many methods utilize human-in-the-loop interactive learning, where humans need to monitor and intervene using shared autonomy \cite{karamcheti2022lila, liu2024sirius} or language corrections \cite{cui2023no, shi2024yell, olaf}. In contrast, we propose an automated failure recovery pipeline that augments existing expert demonstrations into rich language-annotated trajectories, enhancing 3D robotic manipulation and few-shot sim-to-real transfer.

\textbf{Failure Detection and Recovery.} Self-recovery from online failures is crucial for robots. Prior works leverage external failure detectors—either trained models \cite{liu2024sirius, cornelio2024recover} or proprietary LLMs \cite{dai2024think,zheng2024evaluating}—to monitor performance and request human assistance when needed. Other approaches focus on collecting failure recovery trajectories through scalable auto-generation pipelines \cite{hoque2024intervengen, ke2023ccil, ha2023scaling}. For example, I-Gen \cite{hoque2024intervengen} builds on MimicGen \cite{mandlekar2023mimicgen} to automatically generate corrective interventions from a small set of human demos to cover more failed states. Similarly, \cite{ha2023scaling} uses LLMs to verify the robot’s internal information and retry tasks until successful. However, these methods do not incorporate language-guided control, limiting their ability to adapt or shape robot's behavior based on human instructions or linguistic feedback.

\textbf{Language-guided Robot Learning.}
Language is a natural medium for humans to specify tasks and to interact with robots. Most previous works \cite{c2farm, peract, rvt} have focused on using short task goal descriptions to instruct multi-task visuomotor policies. Recent works have been shifting towards enabling real-time human intervention through verbal corrections. For example, OLAF \cite{olaf} uses GPT-4 to re-label incorrect actions based on user feedback like \textit{``move closer to the cup"}. RT-H \cite{belkhale2024rt-h} employs the RT-2 model \cite{brohan2023rt2} to generate language instructions and robot action tokens within predefined hierarchies, allowing for human interventions in a fixed set of spatial movements. Similarly, YAY Robot \cite{shi2024yell} trains a high-level language policy to adjust behaviors on-the-fly by retrieving instructions from a candidate pool and a low-level policy to follow these instructions. However, unlike YAY and other approaches that rely on simple instructions (usually a verb and a noun), our work leverages rich language descriptions, including failure analysis, fine-grained spatial movements, and expected outcomes. Furthermore, our models are trained directly on augmented failure recovery data, reducing the need for additional online data collection to cover failure states and recovery actions.

\section{METHOD}

We develop a data augmentation pipeline to produce rich language-guided failure recovery trajectories and a framework named \modelname, where a VLM (\textit{supervisor}) guides a visuomotor policy (\textit{actor}) for robust robotic manipulation. 

\subsection{Problem Statement}

Our task is language-conditioned robotic control with two sub-problems: (1) language-conditioned multi-task imitation learning for visuomotor policies and (2) single-view image-conditioned language instruction generation for VLMs.  
Assume we have a multi-task dataset of expert demonstrations $\mathcal{D}^\text{expert}$ containing successful trajectories, where each trajectory $\tau=(\delta_1, \delta_2, \cdots, \delta_T, L)$ consists of a sequence of waypoint transitions $\delta_{1:T}$ obtained via heuristic keyframe discovery \cite{c2farm} and a high-level task goal $L$ expressed in natural language like \textit{``close the red jar"}.
Each transition $\delta_t$ is a tuple of ${(o_t, a_t, o_{t+1})}$ at timestep $t$, where $o_t$ is the observation from RGB-D cameras and proprioceptive states and $a_t$ is a 9-dim waypoint action including a 6-DoF EE pose, a binary gripper state, and a binary indicator for planning a collision-free path. A common objective for visuomotor policies is to learn $\pi(a_t|o_t, L)$ from $\mathcal{D}^\text{expert}$.
However, training policies solely on task goal $L$ often leads to overfitting to demonstrations, resulting in poor language understanding and instruction-following \cite{shi2024yell, ha2023scaling, xu2024naturalvlm}.
Therefore, we introduce rich language instruction $\ell_t$ for each keyframe transition to train a better visuomotor policy $\pi(a_t|o_t,\ell_t, L)$. Note that unlike the previous works \cite{shi2024yell, belkhale2024rt-h} that only use simple instructions and require human intervention, we aim to generate more expressive sentences for better language control in an \textbf{automated} way without the necessity of human involvement for failure correction.
In addition, we fine-tune a VLM $p_{\text{vlm}}(\ell_t|o_t, \ell_{t-1}, L)$ for explaining failure states and generating rich language instructions given the current observations, previous instructions and high-level task goals.

\subsection{Data Generation}



\subsubsection{Failure Definition}
 We define failure as a significant deviation from the expert action at a given keyframe, classified into two categories: (1) \textit{recoverable failure}, correctable by using existing expert actions, and (2) \textit{catastrophic failure}, requiring a scene reset due to excessive scene state changes (e.g., objects being knocked over or falling off the table).


\subsubsection{Failure Recovery Augmentation}
we aim to scalably extend existing expert demonstrations with recoverable failures without additional human efforts.
For each trajectory $\tau$ in $\mathcal{D}^\text{expert}$, we identify and perturb a set of \textit{crucial keyframes} that correspond to motion primitives for alignment (e.g., move to the above of objects), grasping (e.g., lower down to pick), and releasing (e.g. place the grasped object down). Heuristic rules based on the gripper's opening state, positional changes, and timestep number are used to determine the crucial keyframes. 
Fig. \ref{fig:data_aug} illustrates the data augmentation strategy in detail.
To be concrete, suppose there is a crucial keyframe at timestep $j$ with an expert transition $\delta_{j-1}={(o_{j-1}, a_{j-1}, o_{j})}$ from the previous timestep.
We then add truncated Gaussian noise to randomly perturb the expert action $a_{j-1}$ into $\tilde{a}_{j-1}=a_{j-1}+\epsilon$ and step through the environment to get the failure transition $\tilde{\delta}_{j-1}={(o_{j-1}, \tilde{a}_{j-1}, \tilde{o}_{j})}$ to cover failure states (see step \circlednum{2} and \circlednum{5} in Fig.\ref{fig:data_aug}). The expert action $a_{j-1}$ can be used as the correction to get recovery transition $\delta^\text{c}_{j}=({\tilde{o}_{j}, a^\text{c}_{j}, o^\text{c}_{j+1})}$, where $a^\text{c}_{j}=a_{j-1}$ and $ o^\text{c}_{j+1}=o_{j}$ (see step \circlednum{3} in Fig. \ref{fig:data_aug}). During our experiment, we found that such \textit{one-step} recovery strategy works for alignment failures, but is not adequate for more complex motion primitives, such as grasping and releasing, as an immediate recovery will lead to catastrophic collision with target objects. Therefore, we also propose a \textit{two-step} recovery strategy (see step \circlednum{6} and \circlednum{7} in Fig.\ref{fig:data_aug}) with an intermediate transition $\delta^\text{i}_{j}=(\tilde{o}_j, a^\text{i}_j, o^\text{i}_{j+1})$ added before the recovery transition $\delta^\text{c}_{j+1}=({o^\text{i}_{j+1}, a^\text{c}_{j+1}, o^\text{c}_{j+2})}$, where $a^\text{c}_{j+1}=a_{j-1}, o^\text{c}_{j+2}=o_{j}$, and $a^\text{i}_j$ is an expert action sampled from the waypoints in between keyframes $j-1$ and $j$. After the augmentation,  all episodes containing failure recovery are rolled out and filtered based on task success. Finally, we obtain a new dataset $\mathcal{D}^\text{recovery}$ that includes additional recovery transitions $\delta^\text{i}$ and $\delta^\text{c}$. 
This procedure significantly offloads human efforts for monitoring the policy online and intervening to correct failures as in \cite{shi2024yell, belkhale2024rt-h}.

\begin{figure*}[t]
    \centering
    \includegraphics[width=0.9\linewidth]{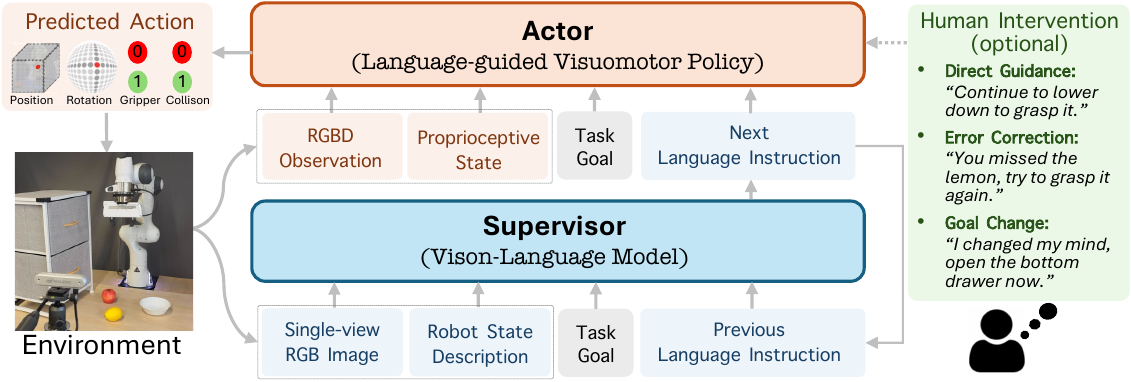}
    \caption{The \modelname\ framework consists of: (1) the \textit{Supervisor}, a VLM that monitors the robot's behavior, providing feedback for task execution and error correction with rich instructions; and (2) the \textit{Actor}, a language-conditioned visuomotor policy that generates actions based on visual observations, proprioceptive states, and language guidance that includes a high-level task goal and an instruction. 
    }
    \label{fig:model2}
    \vspace{-15pt}
\end{figure*}

\subsubsection{Rich Language Annotation}
It is quite challenging for LLMs to generate faithful language descriptions based on numerical action values due to the notorious hallucination problem \cite{xu2024hallucination}. Therefore, we include detailed task descriptions, ground-truth object locations, failure types and heuristic language to construct informative prompts for LLMs. 
Specifically, the heuristic language consists of template-based natural descriptions of the EE pose movement caused by the last action (see the left bottom part in Fig. \ref{fig:data_aug}), in which we compare the state changes between the last and current observations in terms of position, orientation, gripper state, and collision. 
Based on this information, we query GPT-4-turbo to paraphrase the heuristic language into more coherent and expressive natural language.
After $\mathcal{D}^{\textit{recovery}}$ is annotated with rich instructions, we obtain our language-guided failure-recovery dataset  
$\mathcal{D}^{\textit{recovery+lang}}$, where each transition is represented as $\delta_t=(o_t, a_t, \ell_t, o_{t+1})$ and $\ell_t$ refers to the rich  instruction that guides the visuomotor policy. 

\subsection{Model}
\modelname\ consists of two main parts (Fig.~\ref{fig:model2}): (1) a VLM as the supervisor to generate rich language instruction $\ell_t$; and (2) a language-conditioned visuomotor policy as the actor to predict the action $a_t$ to interact with the environment.

\subsubsection{Supervisor} In our experiment, we select the front-view RGB image as the visual input. The robot state description summarizes the change of the robot proprioceptive state after taking the last action. We compare the delta changes in position, rotation, gripper state and collision to generate the description based on predefined templates, e.g., ``\textit{The robot moved (forward$|$backward$|$downward$|$upward$|$left$|$right) (a little$|$significantly)}".
Then, we prepare the input for VLM in a format of   \textit{``$<$image$>$\textbackslash nThe task goal is: \{task\_goal\}. In the previous step, the robot arm was given the following instruction: \{previous\_instruction\}. \{robot\_state\_description\}. Based on the visual observation and the context, how does the robot fulfill that previous instruction and what's the next instruction for the robot?}" to generate the next instruction.

\subsubsection{Actor} The actor is a language-guided visuomotor policy, where any existing waypoint-based models such as RVT \cite{rvt} and PerAct \cite{peract} can be used interchangeably. 
To enhance the capability of these models to understand rich language instructions, we concatenate the high-level task goal $L$ and the rich instruction $\ell_t$ together as the language input for the policy  at each step in the following format as \textit{``Task goal: \{task\_goal\}.\textbackslash nCurrent instruction: \{rich\_instruction\}''}, in which the \{rich\_instruction\} is generated by the VLM. After adapting to our augmented data, visuomotor polices can predict actions based on more descriptive language.

\begin{table*}[t]
\centering
\scalebox{0.96}{
\begin{tabular}{ccccccccccc}
\specialrule{0.3mm}{0mm}{0mm}
Models & \begin{tabular}[c]{@{}c@{}} \texttt{Avg.}\\ \texttt{Succ.} $(\uparrow)$ \end{tabular} & \begin{tabular}[c]{@{}c@{}}\texttt{Avg.}\\ \texttt{Rank} $(\downarrow)$\end{tabular} & \begin{tabular}[c]{@{}c@{}}Close\\ Jar\end{tabular} & \begin{tabular}[c]{@{}c@{}}Drag\\ Stck\end{tabular} & \begin{tabular}[c]{@{}c@{}}Insert\\ Peg\end{tabular} & \begin{tabular}[c]{@{}c@{}}Meat off\\ Grill\end{tabular} & \begin{tabular}[c]{@{}c@{}}Open\\ Drawer\end{tabular} & \begin{tabular}[c]{@{}c@{}}Place\\ Cups\end{tabular} & \begin{tabular}[c]{@{}c@{}}Place\\ Wine\end{tabular} & \begin{tabular}[c]{@{}c@{}}Push\\ Buttons\end{tabular} \\ 
\toprule
PerAct \cite{peract} & 49.4 & 3.7  & 55.2 \text{\tiny ± 4.7} & 89.6 \text{\tiny ± 4.1} & 5.6 \text{\tiny ± 4.1} & 70.4 \text{\tiny ± 2.0} & 88.0 \text{\tiny ± 5.7} & 2.4 \text{\tiny ± 3.2} & 44.8 \text{\tiny ± 7.8} & 92.8 \text{\tiny ± 3.0} \\
RVT \cite{rvt} & 62.9 & 2.2 & 52.0 \text{\tiny ± 2.5} & \textbf{99.2} \text{\tiny ± 1.6}& 11.2 \text{\tiny ± 3.0} & 88.0 \text{\tiny ± 2.5} & 71.2 \text{\tiny ± 6.9} & 4.0 \text{\tiny ± 2.5} & 91.0 \text{\tiny ± 5.2} & \textbf{100.0} \text{\tiny ± 0.0} \\
Act3D \cite{act3d} & 65.0 & 2.2 & \textbf{92.0} & 92.0 & \textbf{27.0} & \textbf{94.0} & 93.0 & 3.0 & 80.0 & 99.0 \\
\modelname\  (Ours)  & \textbf{70.2} \text{\tiny ± 1.13} & \textbf{1.6} & 85.6 \text{\tiny ± 2.0} & \textbf{99.2} \text{\tiny ± 1.6} & 9.6 \text{\tiny ± 4.8} & 91.2 \text{\tiny ± 3.0} & \textbf{100.0} \text{\tiny ± 0.0} & \textbf{6.4} \text{\tiny ± 4.1} & \textbf{98.4} \text{\tiny ± 2.0} &\textbf{ 100.0} \text{\tiny ± 0.0} \\
\specialhdashline
\modelname \text{\scriptsize +H} (Ours)  & 80.1 \text{\tiny ± 0.52}  & -- & 91.2 \text{\tiny ± 1.6} & 100.0 \text{\tiny ± 0.0} & 25.6 \text{\tiny ± 5.4} & 98.4 \text{\tiny ± 2.0} & 100.0 \text{\tiny ± 0.0} & 6.4 \text{\tiny ± 4.1} & 100.0 \text{\tiny ± 0.0} & 100.0 \text{\tiny ± 0.0} \\ 
\specialrule{0.3mm}{0mm}{0mm}
Models & \begin{tabular}[c]{@{}c@{}}Put in\\ Cupboard\end{tabular} & \begin{tabular}[c]{@{}c@{}}Put in\\ Drawer\end{tabular} & \begin{tabular}[c]{@{}c@{}}Put in\\ Safe\end{tabular} & \begin{tabular}[c]{@{}c@{}}Screw\\ Bulb\end{tabular} & \begin{tabular}[c]{@{}c@{}}Slide\\ Block\end{tabular} & \begin{tabular}[c]{@{}c@{}}Sort\\ Shape\end{tabular} & \begin{tabular}[c]{@{}c@{}}Stack\\ Blocks\end{tabular} & \begin{tabular}[c]{@{}c@{}}Stack\\ Cups\end{tabular} & \begin{tabular}[c]{@{}c@{}}Sweep to\\ Destpan\end{tabular} & \begin{tabular}[c]{@{}c@{}}Turn\\ Tap\end{tabular} \\ 
\toprule
PerAct \cite{peract} & 28.0 \text{\tiny ± 4.4} & 51.2 \text{\tiny ± 4.7} & 84.0 \text{\tiny ± 3.6} & 17.6 \text{\tiny ± 2.0} & 74.0 \text{\tiny ± 13.0} & 16.8 \text{\tiny ± 4.7} & 26.4 \text{\tiny ± 3.2} & 2.4 \text{\tiny ± 2.0} & 52.0 \text{\tiny ± 0.0} & 88.0 \text{\tiny ± 4.4} \\
RVT \cite{rvt} & 49.6 \text{\tiny ± 3.2} & 88.0 \text{\tiny ± 5.7} & 91.2 \text{\tiny ± 3.0} & 48.0 \text{\tiny ± 5.7} & 81.6 \text{\tiny ± 5.4} & \textbf{36.0} \text{\tiny ± 2.5} & \textbf{28.8} \text{\tiny ± 3.9} & 26.4 \text{\tiny ± 8.2} & 72.0 \text{\tiny ± 0.0} & 93.6 \text{\tiny ± 4.1} \\
Act3D \cite{act3d} & \textbf{51.0} & 90.0 & \textbf{95.0} & 47.0 & 93.0 & 8.0 & 12.0 & 9.0 & \textbf{92.0} & \textbf{94.0} \\
\modelname\  (Ours)  & 50.4 \text{\tiny ± 4.1} & \textbf{100.0} \text{\tiny ± 0.0} & 93.6 \text{\tiny ± 4.1} & \textbf{72.0} \text{\tiny ± 5.7} & \textbf{99.2} \text{\tiny ± 1.6} & 25.6 \text{\tiny ± 4.1} & 15.2 \text{\tiny ± 3.0} & \textbf{41.6} \text{\tiny ± 5.4} & 84.0 \text{\tiny ± 0.0} & 91.2 \text{\tiny ± 3.0} \\
\specialhdashline
\modelname \text{\scriptsize +H} (Ours)  & 71.2 \text{\tiny ± 6.9} & 100.0 \text{\tiny ± 0.0} & 100.0 \text{\tiny ± 0.0} & 92.0 \text{\tiny ± 0.0} & 100.0 \text{\tiny ± 0.0} & 38.4 \text{\tiny ± 2.0} & 60.0 \text{\tiny ± 5.7} & 69.6 \text{\tiny ± 4.1} & 89.6 \text{\tiny ± 2.0} & 100.0 \text{\tiny ± 0.0} \\ 
\specialrule{0.3mm}{0mm}{0mm}
\end{tabular}
}
\caption{: Multi-task performance comparision of different models on 18 RLbench tasks. \modelname \text{\scriptsize +H} is \modelname\ with human intervention.}
\label{tab: main}
\vspace{-12pt}
\end{table*}

\subsubsection{Online Evaluation}
During evaluation, for each step, we first use the VLM supervisor to analyze current scene (i.e, determine whether the robot started the task or followed the last instruction correctly or made a recoverable failure) and generate a fine-grained instruction, which is then used to instruct the actor to predict an action to control the robot. 


\section{Experiments}






\subsection{Experimental set-up}
We choose RLbench \cite{james2020rlbench} as our benchmark for simulated experiments and test sim-to-real transfer on a Panda robot. 

\subsubsection{Model backbone} 
For the supervisor, we select llama3-llava-next-8B \cite{li2024llavanext-strong}, a latest variant of LLaVA model \cite{llava} due to its superior multimodal reasoning capabilities and the simplicity of fine-tuning under a limited budget. For the actor, we choose RVT \cite{rvt}, one of the state-of-the-art visuomotor policy, and adtrain on our new dataset. 

\begin{table}[t]
    \centering
    \scalebox{1.0}{
    \begin{tabular}{l@{\hskip 4pt}c@{\hskip 6pt}c@{\hskip 6pt}c@{\hskip 6pt}c}
    \specialrule{0.3mm}{0mm}{0mm}
Dataset & Length & \begin{tabular}[c]{@{}c@{}} \# Semantic \\ Roles \end{tabular} & \begin{tabular}[c]{@{}c@{}} \# Unique \\ Tags  \end{tabular} & Example\\
      \toprule
       RT-H \cite{belkhale2024rt-h}  &  4.52 & 1.06 & 2.26 & \textit{``move arm left"} \\
       \specialrule{0.1mm}{0.3mm}{0.3mm}
       YaY \cite{shi2024yell} & 4.73 &  1.04 & 3.79 & \textit{``move to the left"}\\
       \specialrule{0.1mm}{0.3mm}{0.3mm}
       Ours (simple) & 4.38 & 1.06	& 2.69 & \textit{``move left"}\\
       \specialrule{0.1mm}{0mm}{0mm}
       Ours (rich) & 18.28 & 3.64 & 8.31 & 
       \begin{tabular}[c]{@{}c@{}} \textit{``It moved too right,} \\ \textit{correct it position by} \\   \textit{moving slightly left, } \\   \textit{then align with $\cdot$$\cdot$$\cdot$"}\end{tabular} 
\\
       \bottomrule
    \end{tabular}}
    \caption{: Comparison of language richness levels for datasets. 
    }
    \label{tab: lang-comp}
    \vspace{-16pt}
\end{table}

\subsubsection{Augmented Dataset} 
We gather the training and validation expert demos from RLbench as $\mathcal{D}^\text{expert}$ (\num{2250} episodes in total), perturb each episode five times and filter unsuccessful trajectories to obtain $\mathcal{D}^\text{recovery+lang}$ (\num{10159} episodes in total). 
Both simple and rich language instructions are generated by prompting GPT-4-turbo for comparative study. The simple instructions resemble previous works \cite{belkhale2024rt-h, shi2024yell}, consisting of a short sentence that mainly includes a verb (e.g., \textit{move}) and a noun (e.g., \textit{jar}) or direction (e.g., \textit{left}), whereas the rich instructions include more failure explanation, detailed descriptions for the spatial movements, attributes (e.g. color, location and shape) about target objects, and the expected outcome of taking the action. 
Table \ref{tab: lang-comp} compares the richness of language across different datasets, where we measure  the average sentence length, the number of semantic roles \cite{gildea2002automatic}  and unique semantic tags using the AllenNLP toolkit \cite{gardner-etal-2018-allennlp}.


\subsubsection{Training Details} For the supervisor, we use LoRA \cite{hu2021lora} to continually fine-tune the LLaVA model for 2 epochs, with a LoRA rank of 128 and a scaling factor $\alpha$ of 256. 
To stabilize training, we use deepspeed zero2 stage \cite{rajbhandari2020zero} with  a batch size of 64 and a learning rate of 2e-5.
For the actor, we modify RVT by replacing its original language encoder CLIP \cite{radford2021learning} with T5-11B \cite{raffel2020exploring} to enhance its language understanding and removing the timestep input from the proprioceptive state as we find it hinders the language controllability of the policy.
We use the LAMB optimizer \cite{you2019large} to train the modified RVT is for 18 epochs, with a batch size of 48 and a learning rate of 1.5e-3. All training processes take around 30 hours to finish with 8 40GB A40 GPUs.


\subsection{Simulated Experiments}
Following the multi-task learning setup in \cite{peract}, we evaluate our framework over 18 RLbench tasks with a total of 450 episodes. Visual observations are captured from four RGB-D cameras positioned at the front, left shoulder, right shoulder, and wrist. The front-view images are input to the VLM, while all views are input to the visuomotor policy.


\subsubsection{Baselines} 
We compare with the following baselines: 
(1) PerAct \cite{peract} encodes the RGB-D images into a sequence of voxel grid patches and uses the perceiver transformer \cite{jaegle2021perceiver}  to predict actions; 
(2) RVT \cite{rvt} projects the pointcloud into multiple virtual images from orthogonal perspectives and aggregates information across the views via a transformer. As RVT takes images rather than voxels as input, it scales and performs better than PerAct while achieving faster training and inference speed; 
(3) Act3D \cite{act3d} lifts pre-trained 2D CLIP features into 3D using depth sensing, and learns a 3D scene feature field through recurrent coarse-to-fine point sampling and relative-position attention to decode optimal EE actions.


\begin{figure}[t]
    \centering
    \vspace{-3pt}
    \includegraphics[width=0.99\linewidth]{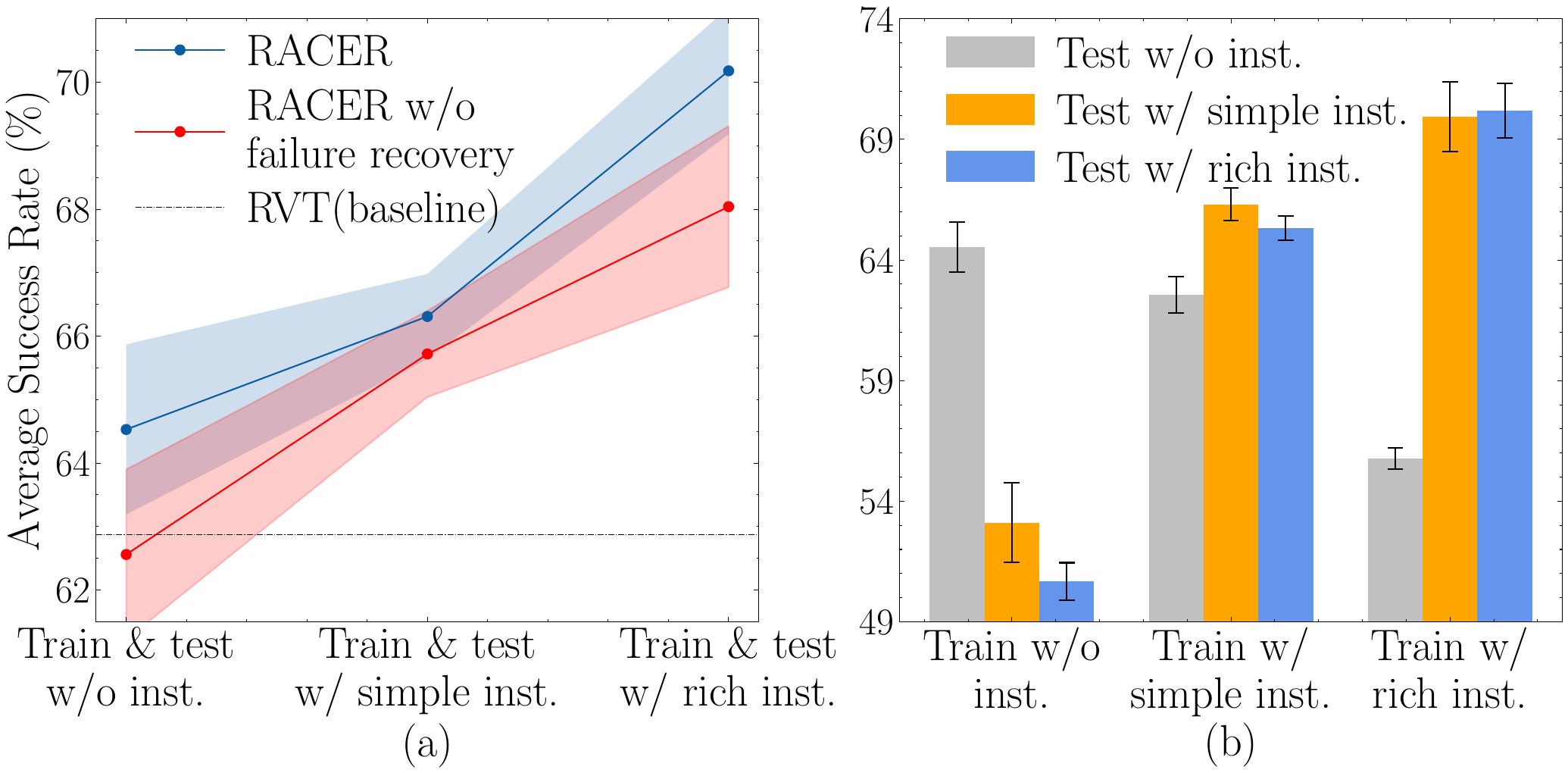}
    \vspace{-17pt}
    \caption{(a) Comparison of \modelname 's performance trained with and without failure recovery across three types of instructions. (b) Cross-evaluation of \modelname\ trained with failure recovery, where training and testing were conducted on varying instruction types.}
    \label{fig:lang-comp}
    \vspace{-18pt}
\end{figure}


\subsubsection{Multi-task Performance} Table \ref{tab: main} summarizes the performance comparisons between \modelname\ trained with rich instructions and the aforementioned baselines. All results are averaged over 5 random seeds on 18 RLbench tasks. When guided by VLM-generated rich instructions, \modelname\ achieves a significant improvement, with an average success rate of 70.2\%. This marks a 7.3\% increase over RVT (62.9\%) and a 5.2\% gain over Act3D (65.0\%). For some long-horizon tasks, such as Put in Drawer, Screw Bulb and Stack Cups, \modelname\ outperforms baselines by 10-24\%. To further comprehensively understand \modelname\ performance, we design experiments to study several questions as follows.

\noindent\noindent\textbf{How does the language richness and failure recovery behaviors improve overall performance?}
To answer this question, we train \modelname s with three types of language input: (1) \textit{no inst.}, i.e., only task goals are provided, no additional instructions; (2) \textit{simple inst.}, i.e., both task goals and simple instructions are given; and (3) \textit{rich inst.}, i.e., both task goals and rich instructions are given. Additionally, we ablate these models by training without any failure recovery transitions. 
For each model, the training and testing conditions are set to be the same.
As shown in Fig. \ref{fig:lang-comp}(a), both failure recovery transitions and rich language inputs substantially improve performance, highlighting their importance for robot manipulation. 
With richer language input, performance consistently increases, and failure recovery data generally boosts performance (around 2\%) across all language settings.

\noindent\textbf{Can \modelname\ trained with rich instructions still perform well given only simple instructions?}
We conduct a cross-evaluation of \modelname s by training and testing them on varied language settings. From Fig. \ref{fig:lang-comp}(b), we surprisingly find that the policy trained with rich instructions is still quite robust to simple instructions during testing, substantially outperforming policies trained and tested both with simple instructions (66.31\%$\rightarrow$69.94\%),  
which underscores the importance and generalizability of rich language training paradigm.
When evaluating without any language instructions, the policy degrades drastically due to the severe mismatch between training and testing conditions. However, it should be noted that our \modelname\ uses a VLM to automatically generate desired instructions without human efforts, thus can circumvent the low performance issue brought by no instruction setting.

\noindent\noindent\noindent\textbf{What is the upper bound performance of \modelname\  when humans can intervene?}
We further conduct human intervention experiments (see \modelname \text{\scriptsize +H} in Tab. \ref{tab: main}) where humans can decide to modify VLM instructions as needed (e.g., when the VLM generates hallucinated or inappropriate instructions) with their own, often \textit{simple}, instructions. This approach increases the success rate from 70.2\% to 80.1\% with only a 24\% intervention rate among the total steps, which demonstrates that our policy, trained on rich instructions, effectively understands and follows unseen human commands as well.

\subsubsection{Task Goal Online Change Experiments} 
We introduce a novel setting to assess the model's robustness by deliberately switching the task goal during execution. For example, in the case of Fig. \ref{fig:data_aug}, we may instruct the robot to place the grasped lid on a different jar just before it is about to release the lid onto the correct jar. The original high-level task goal is replaced with a new one (e.g., \textit{``close the orange jar}"), and a short sentence describing the updated goal is provided to the robot  (e.g., \textit{``No, I changed my mind, move to the orange jar instead."}). After this intervention, no further human instructions are allowed and models need to finish the new task goal on its own.  
We select four tasks from RLbench for testing, each with 25 variations.
As shown in the upper part of Table \ref{tab: combined}, \modelname\ achieves 60 successful episodes out of 100 trials, significantly outperforming RVT, indicating its robustness in handling dynamic goal change scenarios.
\begin{table}[t]
\centering
\scalebox{0.9}{
\begin{tabular}{cccccc}
\specialrule{0.3mm}{0mm}{0mm}
Method & \begin{tabular}[c]{@{}c@{}} \texttt{Avg. Succ.} \\ \textbf{\texttt{(Goal Change)}}\end{tabular}  & \begin{tabular}[c]{@{}c@{}}Close \\ Jar\end{tabular} & \begin{tabular}[c]{@{}c@{}}Screw \\ Bulb \end{tabular} & \begin{tabular}[c]{@{}c@{}} Open \\Drawer  \end{tabular} & \begin{tabular}[c]{@{}c@{}} Push \\ Buttons \end{tabular}\\
\toprule
RVT \cite{rvt} & 9.0 & 0.0 & 20.0 & 16.0 & 0.0 \\
\modelname & \textbf{60.0} & \textbf{64.0} & \textbf{40.0} & \textbf{80.0} & \textbf{56.0} \\
\specialrule{0.3mm}{0mm}{0mm}
Method & \begin{tabular}[c]{@{}c@{}} \texttt{Avg. Succ.} \\ \textbf{\texttt{(Unseen Task)}}\end{tabular} & \begin{tabular}[c]{@{}c@{}}Close \\ Drawer\end{tabular} & \begin{tabular}[c]{@{}c@{}}Move \\ Block \end{tabular} & \begin{tabular}[c]{@{}c@{}} Reach \\ Target  \end{tabular} & \begin{tabular}[c]{@{}c@{}} Pick up \\ Cup \end{tabular} \\
\toprule
RVT \cite{rvt} & 16.0 & 16.0 & 4.0 & 20.0 & 24.0 \\
\modelname & \textbf{47.0} & \textbf{68.0} & \textbf{32.0} & \textbf{40.0} & \textbf{48.0} \\
\toprule
\end{tabular}
}
\caption{: Results for goal change tasks and unseen tasks.}
\vspace{-15pt}
\label{tab: combined}
\end{table}
\subsubsection{Unseen Task Experiments} 
We evaluate our models on unseen tasks to examine \modelname 's zero-shot adaptability. Four new tasks are selected from the RLbench suite, where the objects and manipulation skills may be similar to the training data, but the combinations and scenes are novel. Each task is tested with 25 variations, with results shown in the lower part of Table \ref{tab: combined}. Compared to RVT, \modelname\ performs significantly better, demonstrating its generalizability to generate appropriate instructions based on images from unseen scenarios and execute reasonable actions accordingly.

%


\subsection{Real World Experiments}
We evaluate our model in a real-world setup using a 7-DoF Franka Emika Panda robot and a statically mounted front-view Realsense D455 RGB-D camera. Pointclouds are obtained from the camera as model inputs after extrinsic calibration. 
We choose four tasks (Open Drawer, Place Fruits, Push Buttons, Put Item on Shelf) for testing and collect 60 training demos (15 per task) with failure augmentation (three perturbations each) via manual kinesthetic guidance and GPT-4-turbo language annotation.
Both the visuomotor policy in \modelname\ and the RVT are fine-tuned for 15 epochs with a learning rate of 1.5e-3 and a batch size of 24, while LLaVA is fine-tuned for 2 epochs with a learning rate of 1e-5 and a batch size of 32.  
We compare four ablated models: (1) \modelname$_\textit{scratch}$, trained from scratch on real data; (2) \modelname$_\textit{no inst.}$, trained without instructions on simulated and real data; (3) \modelname$_\textit{simple inst.}$, trained with simple instructions on simulated and real data; and (4) \modelname$_\textit{w/o FA}$, trained with rich instructions but without failure analysis. All models are tested on 40 episodes (10 per task, where 7 are for regular tasks and 3 involves task goal changes).

\begin{table}[t]
\centering
\scalebox{0.9}{
\begin{tabular}{lccccc}
\specialrule{0.3mm}{0mm}{0mm}
Models & \begin{tabular}[c]{@{}c@{}} \texttt{Avg.}\\ \texttt{Succ.} \end{tabular}  & \begin{tabular}[c]{@{}c@{}}Open \\ Drawer\end{tabular} & \begin{tabular}[c]{@{}c@{}} Pick and \\ Place Fruits \end{tabular} & \begin{tabular}[c]{@{}c@{}} Push \\ Buttons  \end{tabular} & \begin{tabular}[c]{@{}c@{}} Put Item \\ on Shelf \end{tabular} \\
\toprule
RVT \cite{rvt} & 25.0 & 10.0 & 30.0 & 20.0 & 40.0 \\
\modelname$_\textit{scratch}$ & 32.5 & 50.0 & 10.0 & 30.0 & 40.0 \\
\modelname$_\textit{no inst.}$ & 25.0 & 60.0 & 0.0 & 0.0 & 40.0 \\
\modelname$_\textit{simple inst.}$ & 32.5 & 60.0 & 10.0 & 20.0 & 40.0 \\
\modelname$_\textit{w/o FA}$ & 62.5 & \textbf{70.0} & 40.0 & 80.0 & 60.0 \\
\modelname & \textbf{72.5} & \textbf{70.0} & \textbf{50.0} & \textbf{100.0} & \textbf{70.0} \\
\bottomrule
\end{tabular}
}
\caption{: Results for real world tasks.}
\vspace{-15pt}
\label{tab: real robot}
\end{table}

As shown in Table \ref{tab: real robot}, \modelname\ demonstrates a significant improvement over RVT, achieving a 47.5\% higher overall success rate.  
This highlights the crucial role of integrating rich descriptions and failure recovery data.
Notably, \modelname\ also substantially outperforms \modelname$_\textit{scratch}$, showing the effectiveness of pre-training in simulation with rich language, which leads to superior sim-to-real transfer. 
When comparing \modelname\ with variants trained with different instruction settings, we observe a steady improvement as the language richness increases, suggesting that more expressive language helps bridge the sim-to-real gap \cite{yu2024natural} for few-shot adaptation.
Additionally, during experiments, we found that RVT and \modelname s trained with simple or no instructions exhibited weaker task understanding, often displaying repetitive behaviors across different tasks (e.g. grasping a drawer handle during the Push Buttons task as timesteps increased), indicating substantial overfitting to the training scenes.
In contrast, training with rich language acts as a form of regularization, preventing overfitting and enabling \modelname\ to achieve more robust control, better failure recovery, and improved adaptation to task-goal changes and scene variations.


\vspace{3pt}
\section{Conclusion and Discussions}
\vspace{3pt}
In this work, we present a scalable language-guided failure recovery data augmentation strategy and introduce \modelname,  a self-recoverable behavior adaptation framework driven by rich language guidance for robotic manipulation.  
Through joint training with rich language instructions and failure recovery data, \modelname\ demonstrates strong performance and robustness across both simulated and real-world environments.
However, our method currently relies on expert demos for data curation and keyframe extraction for sparse waypoint prediction.
In the future, we plan to enhance our data pipeline by augmenting trajectories from human videos and incorporate dense waypoint policies for more precise control. Additionally, we aim to improve \modelname's grounding abilities for better instruction generation and enable it to proactively ask clarifying questions when faced with ambiguity. 
These enhancements will further strengthen \modelname’s  effectiveness and performance in handling complex real-world scenarios.










\clearpage

\bibliographystyle{unsrt}
\bibliography{citation}

\end{document}